\pdfoutput=1
\documentclass[]{article}
\usepackage[letterpaper]{geometry}
\usepackage{amta2022}
\usepackage{times}
\usepackage{url}
\usepackage{latexsym}
\usepackage{layout}

\PassOptionsToPackage{%
round,
semicolon,
authoryear,
}{natbib}
\usepackage[%
  natbibapa
]{apacite}
\usepackage{float}
\usepackage{caption}
\usepackage{subcaption}
\usepackage{multirow}
\PassOptionsToPackage{hyphens}{url}
\usepackage[hang]{footmisc}
\usepackage{hyperref}
\usepackage{booktabs}
\usepackage{footnotehyper}
\usepackage[normalem]{ulem}

\usepackage{mn_styles}
\hypersetup{
    pdftitle={},
    pdfsubject={cs.CL},
    pdfauthor={},
    pdfkeywords={Multi-lingual Neural Machine Translation, Zero-shot, Few-shot, Language tokens, Language tokens},
}

\newcommand{\mST} {\code{S-T}}
\newcommand{\mSTT}{\code{ST-T}}
\newcommand{\mTE} {\code{T-B}}
\newcommand{\mTT} {\code{T-T}}
\newcommand{\EngData}{\mathit{{E}\!\mathbin{\Leftrightarrow}\!{X}}}
\newcommand{\DirectData}{\mathit{{X}\!\mathbin{\Leftrightarrow}\!{Y}}}

\title{
    \bf Language Tokens: \\
    A Frustratingly Simple Approach
    Improves Zero-Shot Performance
    of Multilingual Translation

}

\author{
    \name{\bf Muhammad ElNokrashy} \hfill
        \addr{{muelnokr@microsoft.com}} \\
    \name{\bf Amr Hendy} \hfill
        \addr{{amrhendy@microsoft.com}} \\
    \name{\bf Mohamed Maher} \hfill
        \addr{{mohamedmaher@microsoft.com}} \\
    \name{\bf Mohamed Afify} \hfill
        \addr{{mafify@microsoft.com}} \\
    \addr{Microsoft ATL, Cairo}
\AND
    \name{\bf Hany Hassan Awadalla} \hfill
        \addr{{hanyh@microsoft.com}} \\
      \addr{Microsoft, Redmond}
}

\begin{document}

\maketitle
\pagestyle{empty}

\begin{abstract}
    This paper proposes a simple yet effective method to improve direct (\emph{X-to-Y}) translation for both cases: zero-shot and when direct data is available. We modify the input tokens at both the encoder and decoder to include signals for the source and target languages. We show a performance gain when training from scratch, or finetuning a pretrained model with the proposed setup.
    In the experiments, our method shows nearly $10.0$ BLEU points gain on in-house datasets depending on the checkpoint selection criteria.
    In a WMT evaluation campaign,
    \textit{From-English} performance improves by $4.17$ and $2.87$ BLEU points, in the zero-shot setting, and when direct data is available for training, respectively. While \emph{X-to-Y} improves by $1.29$ BLEU over the zero-shot baseline, and $0.44$ over the many-to-many baseline.
    In the low-resource setting, we see a $1.5\sim\!1.7$ point improvement when finetuning on \emph{X-to-Y} domain data.
\end{abstract}


\section{Introduction}
\label{sec:intro}
Neural machine translation (\newterm{NMT}) has witnessed significant advances since the introduction of the transformer model \citep{vaswani2017attention}. This model has shown impressive performance for bilingual translation commonly from and to English \citep{hassan2018achieving}. It has also been shown that the proposed model could be easily extended to multiple language pairs \citep{johnson-etal-2017-googles,aharoni-etal-2019-massively,wang-etal-2020-balancing,fan2020beyond}, to and/or from English, by simple modifications to the basic architecture. This holds promise for improved performance for low-resource pairs through transfer learning, as well as better training and deployment costs per language pair. This setting is referred to as multilingual neural machine translation (\newterm{MNMT}).

The mainstream method of training \oldterm{MNMT} is to introduce an additional input tag at the encoder to indicate the target language, while the decoder uses the usual begin-of-sentence (\code{BOS}) token. This simple modification to the bilingual architecture is shown to work well up to hundreds of language pairs \citep{Tran2021FBWMT,fan2020beyond}, given a corresponding increase in the number of parameters to handle the increased training data. Despite the emergence of modified architectures which add language-specific parameters, like language specific sub-networks (LASS) \citep{lin-etal-2021-learning}, and adapters \citep{bapna-firat-2019-simple}, the basic architecture remains the most effective choice for deploying large scale production systems.

\section{Motivation}
\label{sec:motivation}
While MNMT was originally focused on English-centric translation, there is increasing interest interest in direct translation%
    \footnote{Also known as \emph{X-Y} translation. In the rest of the paper we refer to translation between any two language pairs not involving English as \emph{direct} or \emph{X-Y} translation.}
rather than pivoting through a common language (ex. English). In \cite{freitag2020complete}, the authors mine direct translation data by matching the English part of English-centric corpora, then use modified temperature sampling to alleviate the over-representation of English as target. Another work \citep{fan2020beyond} leverages public direct translation data including \dataset{CCMatrix} \citep{schwenk2019ccmatrix} and \dataset{CCAligned} \citep{elkishky2019ccaligned} as well as improved sampling and sparse modeling to develop a 100-by-100 direct translation model.

While the availability of direct translation training data helps improve the corresponding directions, the zero-shot case remains of particular importance considering the difficulty in maintaining coverage of the set of rapidly increasing directions, and handling the corresponding increase in data size and compute time and resources. Hence the interest in techniques that improve zero-shot translation performance, benefit from parallel training data as it becomes available,
and which can be easily applied to the basic architecture and pretrained models.

\paragraph{Related Works.}
\citet{yang2021improving} approaches the off-target translation problem using gradient projection and no direct training data. \citet{zhang2020improving} improves the zero-shot case using online back-translation and by specializing layers (ex. LayerNorm) for the target language. \citet{rios-etal-2020-subword} utilizes separately-trained vocabularies per language. \citet{Arivazhagan2019Missing} proposes an alignment loss to enforce source language invariance. \citet{ha-etal-2016-toward} proposes tagging input tokens by the source language and indicating the target language directly to a shared decoder.

\paragraph{Proposal.}
\begin{figure}[!t]
    \centering
    \includegraphics[width=0.99\textwidth]{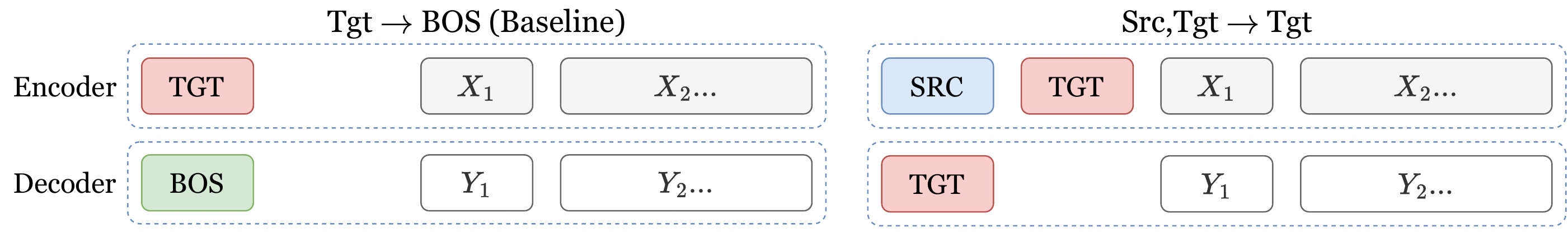}
    \caption{Comparing tokens as seen by the Encoder and the Decoder in the \figleft{} baseline (\mTE{}) and in the \figright{} top proposed method (\mSTT{}).}
    \label{fig:diagram}
\end{figure}
We propose a simple yet effective method that improves the performance of direct translation: The input tokens used in \oldterm{MNMT} are changed to \mSTT{} instead of \mTE{} (see \Figref{fig:diagram}). The encoder takes tokens for both the source and target languages (\code{S,T}) while the decoder takes one for only the target language (\code{T}). It is shown that using these modified tokens significantly improves the performance on direct translation pairs without any parallel $\DirectData{}$ translation data---training only on English-centric data ($\EngData{}$). Remarkably, these gains are quickly obtained if we start from a model trained using the baseline tokens and continue training after adding the new tokens. In subsequent experiments, we also show that some gains are still observed if we continue training the baseline model using a mix of direct ($\DirectData{}$) and English-centric training data---suggesting the method extends to the non zero-shot case as well.
%

The paper is organized as follows: We describe the proposed method in \Secref{sec:Methods}. This is followed by describing the data and the models used in our experiments in \Secref{sec:Data} and \Secref{sec:Models} respectively. \Secref{sec:results-main} gives the experimental results (domain finetuning results in \Secref{sec:Domain-exp}).
Finally, \Secref{sec:conclusion} gives the conclusion.  

\section{Approach}
\label{sec:Methods}

In basic \oldterm{MNMT}, the source sentence is followed by a token indicating the target language at the encoder side, and with the begin-of-sentence token at the decoder side. This setup is \mTE{} in \BFigref{fig:diagram}.
In the proposed method, we perform a simple modification by adding both the source and target tokens to the input at the encoder side\footnote{Note that the order of source and target is not significant and that we also add the target at the decoder.}, and the target at the decoder side. This setup is \mSTT{} in \Figref{fig:diagram}.
%
\cite{wang-etal-2018-three} shows that adding the \code{TGT} language to the decoder input helps English-to-X translation.
In a recent submission to WMT21, \cite{Tran2021FBWMT} uses a \code{SRC} token at the encoder and a \code{TGT} token at the decoder, which can be observed from the public evaluation code\footnotemark{}. In initial experiments we try several variants of indicating the languages to the model. We find that most are similar for the English-centric case, but the proposed method (\mSTT{}) performs the best for zero-shot direct translation.
\footnotetext{
    Found at: \href{https://github.com/facebookresearch/fairseq/blob/47c58f0858b5484a18f39549845790267cffee1a/examples/wmt21/eval.sh}{\nolinkurl{https://github.com/facebookresearch/fairseq/blob}}
\\\hphantom{Found at:\ \ }
    \href{https://github.com/facebookresearch/fairseq/blob/47c58f0858b5484a18f39549845790267cffee1a/examples/wmt21/eval.sh}{\nolinkurl{47c58f0858b5484a18f39549845790267cffee1a/examples/wmt21/eval.sh}}
}
%
\subsection{Initial Experiments}
\label{sec:initial}
\begin{figure}[t]
\centering
    \begin{subfigure}[b]{0.5\textwidth}
        \centering
        \includegraphics[width=\textwidth]{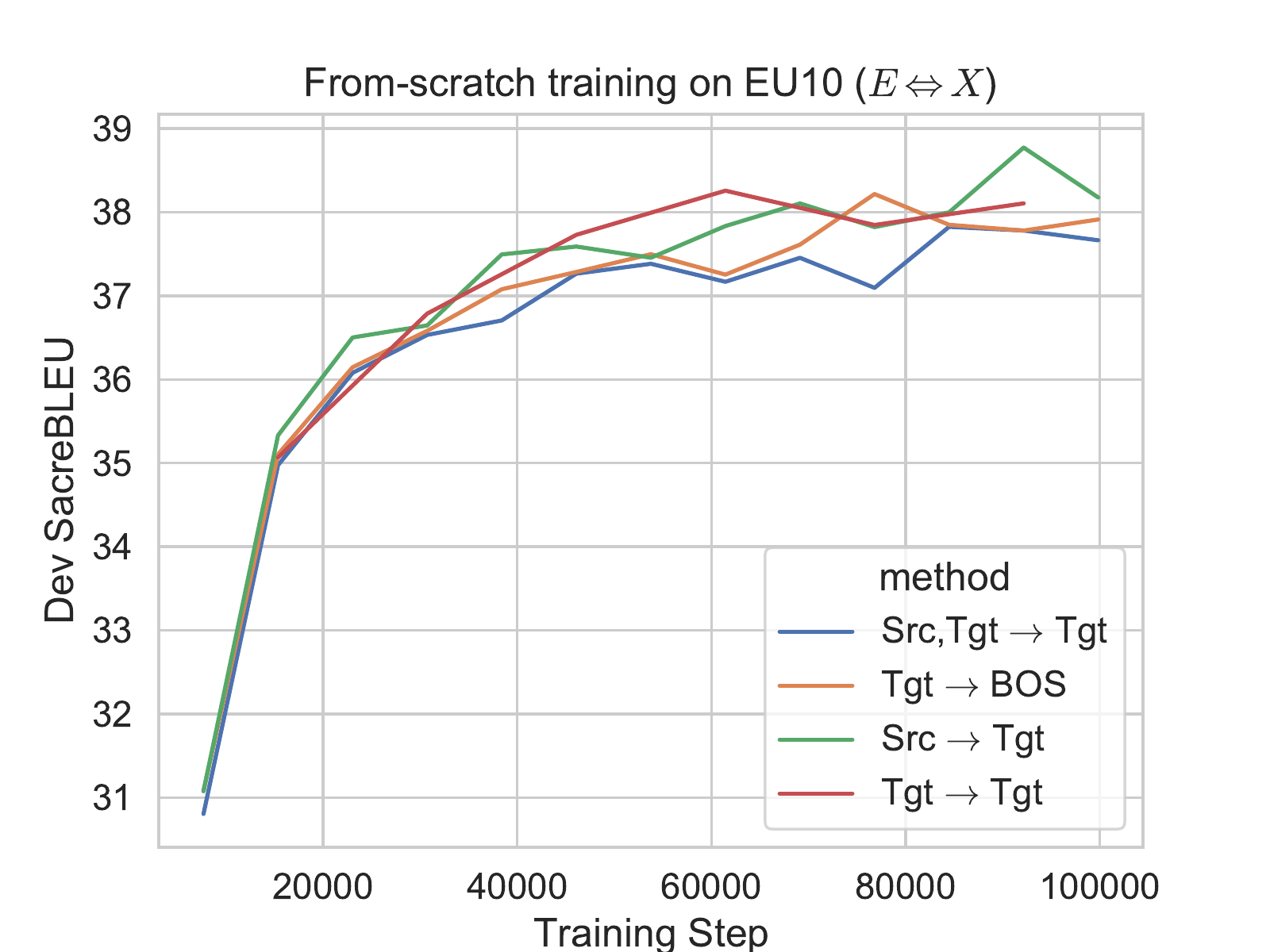}
        \caption{SacreBLEU for $\EngData{}$ dev.}
        \label{fig:eu10_ex_scratch}
    \end{subfigure}%
    \begin{subfigure}[b]{0.5\textwidth}
        \centering
        \includegraphics[width=\textwidth]{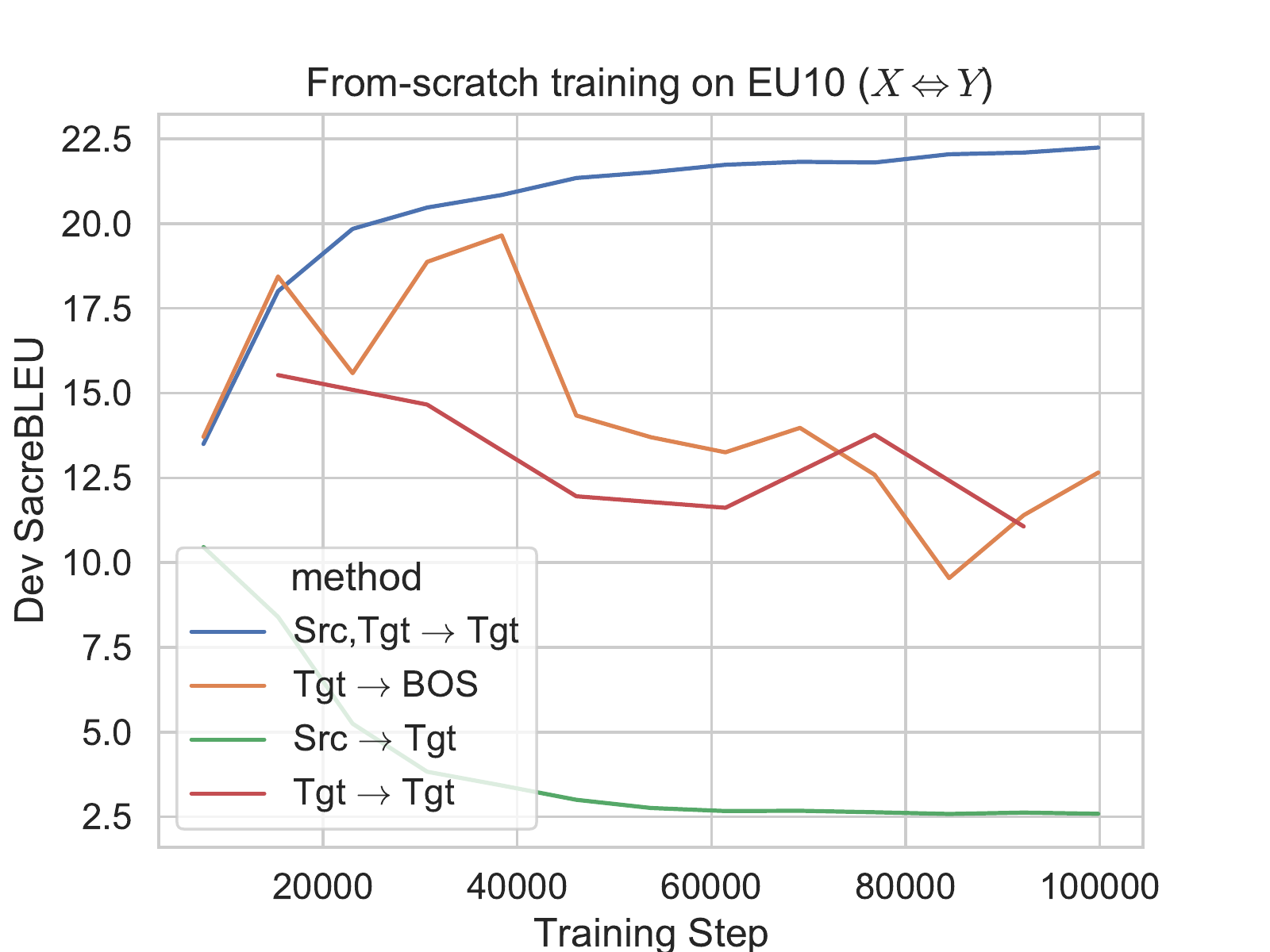}
        \caption{SacreBLEU for $\DirectData{}$ dev.}
        \label{fig:eu10_xy_scratch}
    \end{subfigure}
\caption{
    \textbf{Validation for from-scratch training.}
    All training uses English-centric data and no direct ($\DirectData{}$) data.
    The baseline (\mTE{}) quickly loses performance on $\DirectData$ dev. The \mST{} method fails on $\DirectData{}$ dev, but matches or exceeds alternatives on $\EngData{}$ dev. The proposed method (\mSTT{}) matches the alternatives on $\EngData$ dev (within $1.0$ BLEU) and maintains high performance for $\DirectData{}$ generalization.
}
\label{fig:eu10_scratch}
\end{figure}

To validate the proposed method, we train a model on 10 European languages using in-house English-centric data---Once using the baseline tokens, and once using the new tokens. Details of the model and the training are given in \Secref{sec:Models}.
The graph of the dev BLEU score during training is shown for the English-centric devset and the direct devset in \textbf{Figures \ref{fig:eu10_ex_scratch}} and \textbf{\ref{fig:eu10_xy_scratch}}. Also shown in the figure is the \mST{} setup similar to \citep{Tran2021FBWMT}, and the \mTT{} setup which passes the target language to both encoder and decoder.
%
\subsection{Language Coding and Model Conditioning}
While all models perform similarly for the English-centric set on which they are trained, the behavior is different for the novel direct set. The baseline and proposed models are close at the beginning of the training but the baseline quickly deteriorates \emph{as it improves in its assigned task} on English-centric data.
One explanation is that the conditioning in the \mTE{} case explicitly indicates only the target language, while the source language is inferred. 
Consider also the absence of direct $\DirectData{}$ training data, then an \textit{implicit, and valid, pattern} emerges: When \code{TGT$\ne$en}, it is implied that \code{SRC$=$en}.
Thus, in the $\DirectData{}$ test case, the model expects the source to be in English, which may be a source of confusion.
Conversely, the model with the \mST{} setup performs very poorly from the beginning for the direct set. We propose that this is in line with findings that show that encoder capacity may be of higher importance to \oldterm{MT} than decoder capacity \citep{kim2019research,kasai2020deep}. Removing the \code{TGT} signal from the encoder would then be a significant handicap. It may work in the English-centric case because the \textit{implicit pattern} described above is sufficient conditioning.
\paragraph{Off-Target Translations.}
\BTableref{tab:eu10.lid} shows Language ID mis-matches for \mTE{} and \mSTT{} using fasttext language identification on the \textit{from-scratch} experiments on EU10 (\figref{fig:eu10_scratch}).
\begin{table}[t]
\centering
\begin{tabular}{@{}ccc@{}}
\toprule
\multicolumn{1}{l}{\textbf{Directions}}
& \multicolumn{1}{c}{{\mTE{}}}
& \multicolumn{1}{c}{{\mSTT{}}} \\ \midrule
$\EngData{}$     & $ 0.14\%$   & $0.15\%$ \\
$\DirectData{}$  & $29.36\%$   & $1.69\%$ \\
\midrule
\textbf{Both} & $23.49\%$ & $\mathbf{1.35\%}$ \\
\bottomrule
\end{tabular}
\caption{Percentage of \emph{off-target} samples in the $\EngData{}$ and $\DirectData{}$ dev sets for EU10 measured at $53$k steps. Training graphs in \Figref{fig:eu10_scratch}.}
\label{tab:eu10.lid}
\end{table}
\subsection{Building on Pretrained Models}
\begin{figure}[h]
\centering
    \begin{subfigure}[b]{0.5\textwidth}
        \centering
        \includegraphics[width=\textwidth]{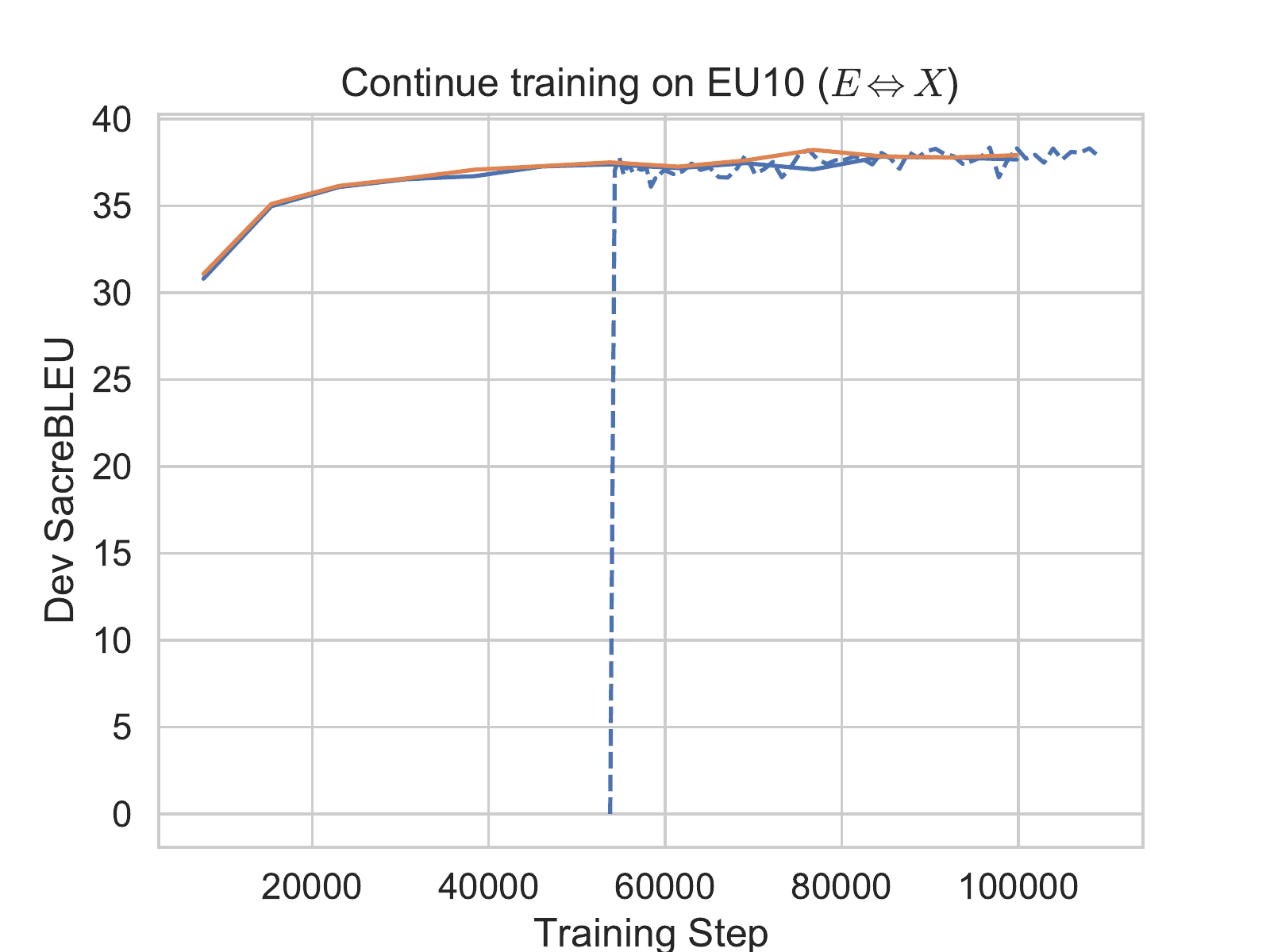}
        \caption{SacreBLEU for $\EngData{}$ dev.}
        \label{fig:eu10_ex_cont}
    \end{subfigure}%
    \begin{subfigure}[b]{0.5\textwidth}
        \centering
        \includegraphics[width=\textwidth]{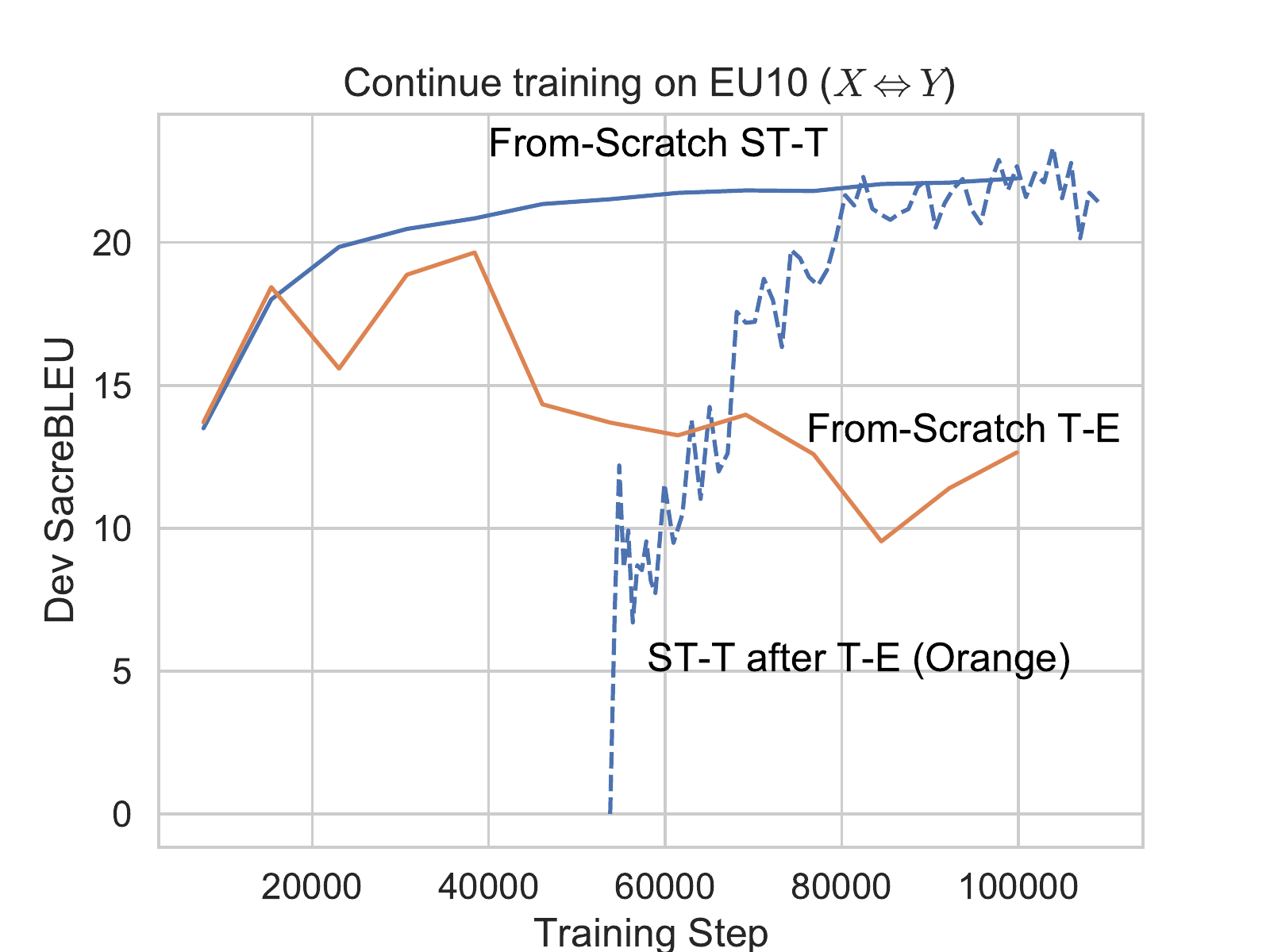}
        \caption{SacreBLEU for $\DirectData{}$ dev.}
        \label{fig:eu10_xy_cont}
    \end{subfigure}
\caption{
\textcolor{orange}{Orange} is the baseline (\mTE{}) setup, while \textcolor{blue}{blue} is the proposed (\mSTT{}) setup. The \uline{solid} lines start from scratch. The \textcolor{blue}{\dashuline{dashed}} line continues training from step $53$k of the baseline, using \mSTT{} tokens, on English-centric ($\EngData{}$) data.
No direct $\DirectData{}$ data is used for training.
$\EngData{}$ performance is quickly regained.
$\DirectData{}$ performance approaches that of from-scratch training within a similar \emph{total} training budget (\dashuline{dashed} vs. \uline{solid} \textcolor{blue}{blue} lines).
}
\label{fig:inhouse_continue}
\end{figure}
An interesting scenario is how to make use of already trained models that would be costly to retrain.
We validate our top proposed method in that case by continuing training from a midway checkpoint of the baseline. The performance on both dev sets is shown in \BFigref{fig:inhouse_continue}.
Performance (\dashuline{dashed} {\color{blue} blue} line) on the $\EngData{}$ set starts at zero but rapidly regains its baseline value, then remains steady as training progresses. The same happens for $\DirectData{}$ data but at a slower pace.
The weights already trained on English-centric data seem to realign to the new setup efficiently.

To summarize: Compared to the baseline, the proposed tokens perform similarly on English-centric tests and significantly outperform on direct translation tests while training using English-centric data in both cases. If we continue training the baseline model by adding the proposed tokens we quickly recover the performance of both the English-centric and direct data to levels obtained by training from scratch, as seen in the initial experiments. Therefore, in the rest of this paper we focus on continuing training a model pretrained using the baseline tokens.

\section{Data}
\label{sec:Data}
In initial experiments we build a model for 10 European languages using in-house data (\Secref{sec:Data.eu10}). Follow-up experiments use WMT data \citep{akhbardeh-etal-2021-findings} as well as other publicly available data covering 6 languages (\Secref{sec:Data.wmt}).
Most experiments use English-centric training data, some use direct training data (\Secref{sec:Data.xy}), and some use domain data (\Secref{sec:Data.domain}).
Validation data is described in \Secref{sec:Data.devtest}. Tables are in Appendix \ref{app:data_details}.

\subsection{EU10 Training Data}
\label{sec:Data.eu10}
EU10 is an in-house web-crawled parallel dataset with a total of $3.35$ billion sentence pairs covering 10 European languages: Dutch (nl), English (en), French (fr), German (de), Greek (el), Italian (it), Polish (pl), Portuguese (pt), Spanish (es), and Romanian (ro). Details in \Tableref{tab:eu10.paralleldata}.

\subsection{WMT Training Data}
\label{sec:Data.wmt}
For WMT, we use the data for 12 English-centric language pairs provided by the news translation shared task in WMT21\footnote{\url{https://www.statmt.org/wmt21/translation-task.html} \citep{akhbardeh-etal-2021-findings}.}, and additionally data from the public sources \dataset{CCMatrix} \citep{schwenk2019ccmatrix} and \dataset{CCAligned} \citep{elkishky2019ccaligned}.
The combined set covers the directions of English (en) to and from: Czech (cs), German (de), Icelandic (is), Japanese (ja), Russian (ru), and Chinese (zh).
We apply some preprocessing steps to filter noisy data. We filter for the expected languages using fasttext \citep{joulin-etal-2017-bag}; normalize punctuation using moses\footnote{\url{https://github.com/moses-smt/mosesdecoder} \citep{koehn-etal-2007-moses}.}; then discard sentences longer than $250$ words or with a source/target or target/source length ratio exceeding $3$. The filtered data totals $2.16$ billion sentence pairs. Details on filtered parallel data sizes are shown in \Tableref{tab:wmt.paralleldata}. Note the difference in the parallel data of English-centric directions for the same non-English language is due to having different amounts of synthetic data released by the WMT21 shared task.

\subsection{Direct Training Data}
\label{sec:Data.xy}
For experiments with direct $\DirectData{}$ data covering the $7$ languages in the WMT dataset (\Secref{sec:Data.wmt}), we build a dataset of parallel training data in $42$ translation directions, including the English-centric directions. We collect $\DirectData{}$ data from publicly available sources as described in \Secref{sec:Data.wmt}. The resulting size of the collected bitext $\DirectData{}$ data is shown in \Tableref{tab:xy.paralleldata}. We then sample $10$ million sentence pairs from the WMT English-centric dataset for each direction to avoid catastrophic forgetting on $\EngData{}$ directions. We end up with a many-to-many dataset with a total of $525$ millions sentence pairs that contains $405$ millions sentence pairs in $\DirectData{}$ directions, and $120$ millions sentence pairs in English-centric directions.

\subsection{Development and Test Data}
\label{sec:Data.devtest}
We evaluate the translation performance on various devsets depending on the training dataset and the language list. For experiments using in-house training data (\Secref{sec:Data.eu10}), we use in-house dev sets covering all the many-to-many 90 translation directions. For experiments using the WMT data described in \Secref{sec:Data.wmt} and \ref{sec:Data.xy}, we use publicly available dev sets composed of: \dataset{WMT21}-provided sets to cover the 12 English-centric directions, and the \dataset{Flores-101} benchmark \citep{goyal2021flores101} to cover the remaining 30 $\DirectData{}$ directions.

\subsection{Data for Domain Experiments}
\label{sec:Data.domain}
We utilize domain data from the \dataset{OPUS} project\footnote{\url{https://opus.nlpl.eu/index.php} \citep{tiedemann-2012-parallel}.}. We collect the \dataset{EMEA}, \dataset{JRC-acquis} and \dataset{Tanzil} domains in the directions German to/from Czech. The data is shuffled and split into train, test and validation sets. Any sentences that occur in the validation or test sets are removed from training. The sizes of the splits are shown in \Tableref{domain-data-size}.
\dataset{EMEA} is a parallel corpus of PDF documents from the European Medicines Agency.
\dataset{JRC-Acquis} is a collection of legislative text of the European Union that comprises selected texts written between the 1950s and now.
\dataset{Tanzil} is a collection of Quran translations compiled by the Tanzil project.

\section{Models}
\label{sec:Models}
All experiments use the same architecture and configuration. We use the Transformer encoder-decoder architecture \citep{vaswani2017attention} as the base model and opt for a deep encoder and a shallower decoder as presented in \cite{kim2019research} and \cite{kasai2020deep}, with $24$ encoder layers and $12$ decoder layers. Dimensions are $1024$ for model width, $4096$ for the feed-forward hidden layer, and $16$ attention heads. We use pre-layer normalization which is becoming more common for similar architectures \citep{xiong2020layer}. We use a vocabulary of size $128,000$ with the sentencepiece tokenizer\footnote{\url{https://github.com/google/sentencepiece} \citep{kudo-richardson-2018-sentencepiece}.}. The model size is $0.6$B parameters.
All models are trained by the RAdam optimizer \citep{radam}.
See Appendix \ref{app:Models.Hyper}, \Tableref{tab:hyper} for other hyper-parameters.

\section{Experimental Results}
\label{sec:results-main}

In this section, we show experimental results using the WMT model as described in \Secref{sec:Models}. We first continue training the WMT model using the proposed tokens on English-centric data only, then we continue training using direct data%
\footnote{This is a mix of $\DirectData{}$ and $\EngData{}$ data (\Secref{sec:Data.xy}) to avoid catastrophic forgetting of English-centric translation.}.
In other experiments, we use $\DirectData$ data from start, once with each of \mTE{} and \mSTT{} tokens.
In all cases, we report the BLEU score for both the direct and English-centric dev sets. See \Tableref{tab:results.wmt.bleu}.
The results of continuing training the baseline tokens with direct data are shown in the fourth row (\code{D} \modelname{Direct FT}).
Note that the third row (\code{P-D} \modelname{Proposed $\hookrightarrow$ Direct FT}) corresponds to continuing training with new tokens for 47k iterations, then adding the direct data---running for 112k steps in total.

\subsection{Medium-resource MNMT}
In \BTableref{tab:results.wmt.bleu}, the first row shows the scores of the base WMT model on both English-centric and direct dev sets (in zero-shot setting) with the \mTE{} setup. Continuing to train the model using the new tokens shows gains on both dev sets, although less than what would be expected from the initial results on EU10 (\Secref{sec:initial}). Continuing to train the base model using direct data shows larger gains on the direct dev set, but a smaller gain on English-centric dev.

\begin{figure}[!ht]
\centering
    \begin{subfigure}[b]{0.5\textwidth}
        \centering
        \includegraphics[width=\textwidth]{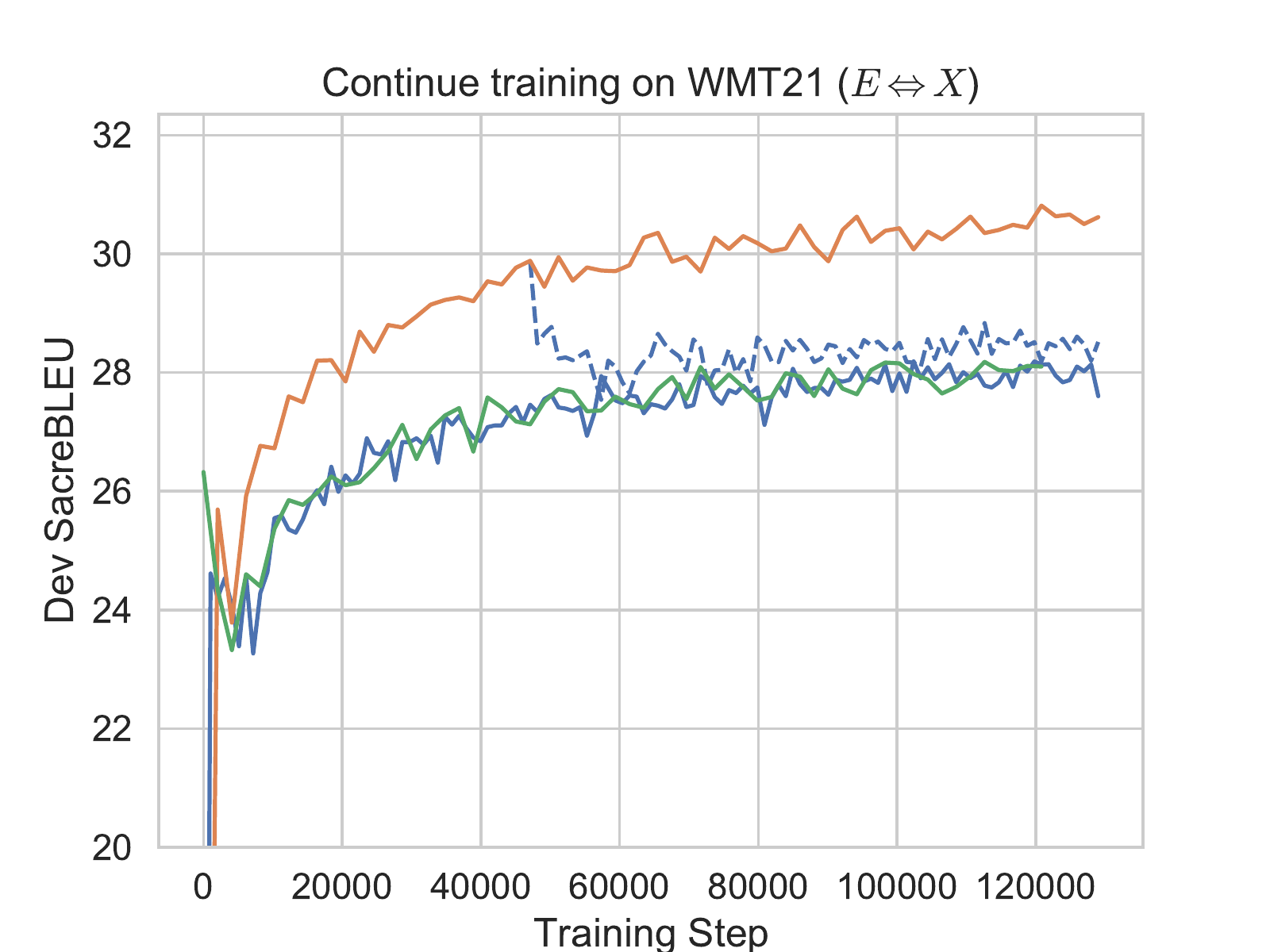}
        \caption{SacreBLEU for $\EngData{}$ dev.}
        \label{fig:wmt_ex_cont}
    \end{subfigure}%
    \begin{subfigure}[b]{0.5\textwidth}
        \centering
        \includegraphics[width=\textwidth]{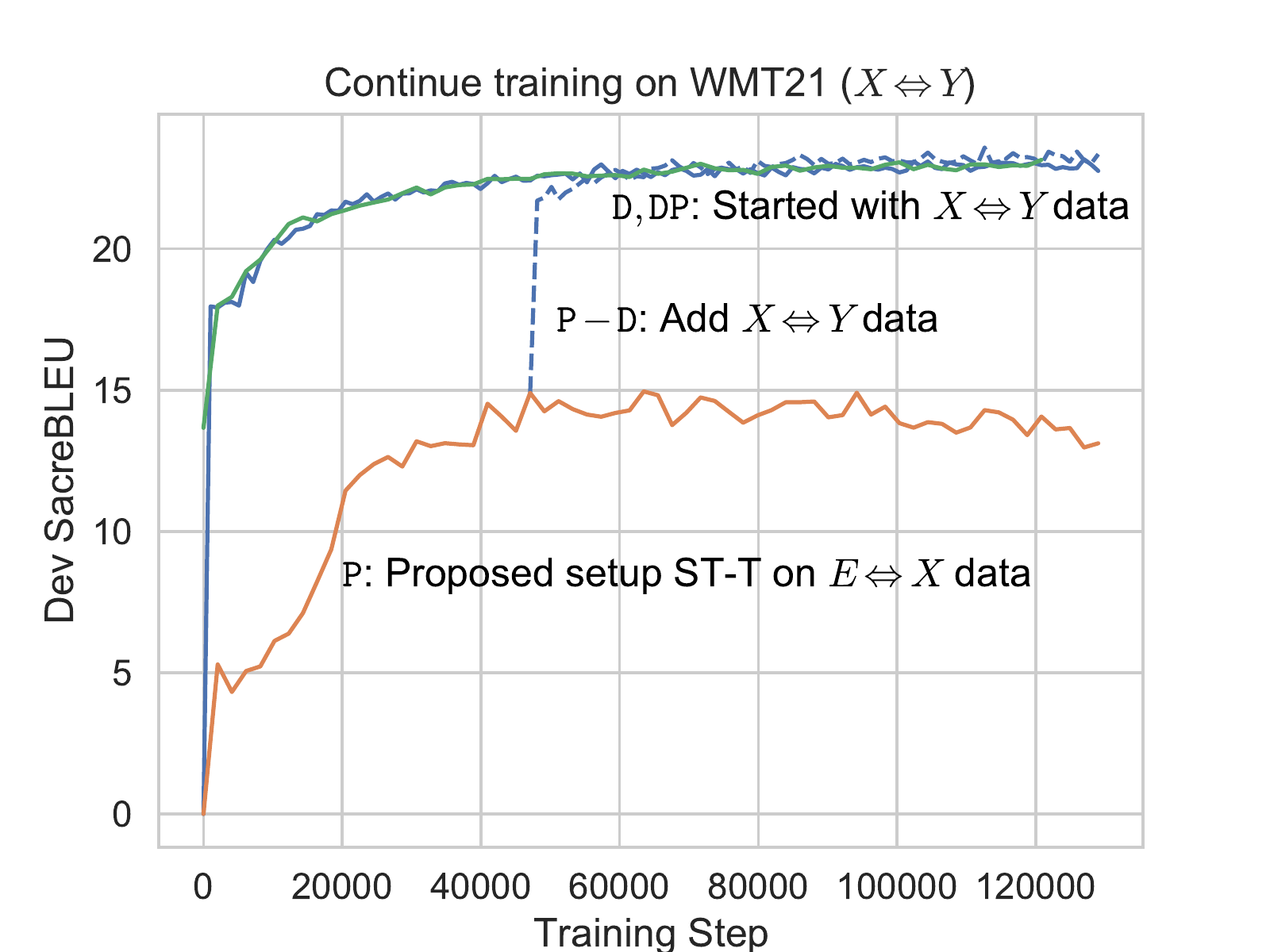}
        \caption{SacreBLEU for $\DirectData{}$ dev.}
        \label{fig:wmt_xy_cont}
    \end{subfigure}
\caption{
All \uline{solid} line models start from the \textbf{pretrained base model} (row \code{B} in Tables \ref{tab:results.wmt.bleu} \& \ref{tab:results.wmt.comet}).
\textcolor{orange}{Orange} is the proposed setup trained on $\EngData{}$ data (row \code{P}).
\textcolor{blue}{Blue} adds $\DirectData{}$ data (rows \code{P-D} \& \code{DP}) where:
Run \code{P-D} (\dashuline{dashed} line) continues training on $\DirectData{}$ data from step $47$k of \textcolor{black}{run} \code{P}, while run \code{DP} (solid blue) starts with that data.
Note that \code{D} and \code{DP} perform similarly on both dev sets, but \code{P-D} improves $\DirectData{}$ performance while lessening the loss of $\EngData{}$ performance that is gained from {\color{orange} run} \code{P}.
\code{P-D} reaches similar performance to \code{D} and \code{DP} in fewer steps with access to $\DirectData{}$ data directly, suggesting improved data efficiency.
}
\label{fig:wmt_continue}
\end{figure}

The third and last rows show that we still observe some gain from the new tokens after adding the direct data. The best strategy (row 3) is two phases: to train first using the English-centric data, then add the direct data. Consider \Figref{fig:wmt_xy_cont}: It may require a smaller $\DirectData{}$ dataset (fewer steps) in the two-phase setup than when using direct data from the start.
\begin{table}[h]
\centering
\begin{tabular}{@{}l l c c | c c c@{}}
\toprule
    \# &
    Setup & \textbf{Direct} & English-Centric & $X\!\Rightarrow\!E$ & $E\!\Rightarrow\!X$ & Best At \\
\midrule
    B & Base Model                       & 13.67 & 26.30 & 27.37 & 25.28 & - \\ 
    P & $\hookrightarrow$ Proposed           & 14.96 & \textbf{30.27}
                                         & \textbf{31.10} & \textbf{29.45} & 63k \\ 
    P-D & \quad $\hookrightarrow$ Direct FT (from 47k)
                                         & \textbf{23.59} & \textit{28.83}
                                         & \textit{29.52} & \textit{28.15}
                                         & 112k \\ 
    D & $\hookrightarrow$ Direct FT          & 23.15 & 28.10 & 28.90 & 27.30 & 120k  \\ 
    DP & $\hookrightarrow$ Proposed \& Direct FT
                                         & 23.09 & 28.19 & 28.85 & 27.52 & 102k \\ 
\bottomrule
\end{tabular}
\caption{
\textbf{SacreBLEU}\protect\footnotemark{} of the base \dataset{WMT} model (first row) when finetuned in various setups. A hooked arrow ($\hookrightarrow$) indicates a row that continues training from the parent model. Thus rows \code{P}, \code{D}, and \code{DP} show finetuning using the proposed (\mSTT{}) tokens, direct data, or both; while row \code{P-D} continues from row \code{P}. \modelname{Direct FT} refers to finetuning with direct data. The \textit{\textbf{Direct}} column uses \dataset{Flores-100} dev, while the \textit{English-centric} column uses \dataset{WMT21} dev. The \textit{Best~At} column reports the \emph{total} training steps starting from \modelname{Base Model}.}
\label{tab:results.wmt.bleu}
\end{table}
\footnotetext{SacreBLEU signature: \texttt{nrefs:1|case:mixed|eff:no|tok:13a|smooth:exp|version:2.1.0}. For targets in Chinese and Japanese, the tokenizers used are \code{zh} and \code{ja-mecab}. \citep{post-2018-call}}

\begin{table}[h]
\centering
\begin{tabular}{@{}l l c c | c c c@{}}
\toprule
    \# &
    Setup & \textbf{Direct} & English-Centric & $X\!\Rightarrow\!E$ & $E\!\Rightarrow\!X$ & Best At \\
\midrule
    B & Base Model                       & -16.23 & 36.00 & 32.69 & 39.31 & - \\ 
    P & $\hookrightarrow$ Proposed           & 6.36 & \textbf{49.87} & \textbf{45.21} & \textbf{54.54} & 63k \\ 
    P-D & \quad $\hookrightarrow$ Direct FT (from 47k)
                                         & \textbf{54.30} & \textit{47.14}
                                         & \textit{44.18} & \textit{50.10}
                                         & 112k \\ 
    D & $\hookrightarrow$ Direct FT          & 51.74 & 44.71 & 41.73 & 47.69 & 120k  \\ 
    DP & $\hookrightarrow$ Proposed \& Direct FT
                                         & 51.57 & 44.47 & 41.82 & 47.12 & 102k \\ 
\bottomrule
\end{tabular}
\caption{
Average \textbf{COMET}\protect\footnotemark{} ($\times 100$) of the set of WMT experiments.
Same notation as \Tableref{tab:results.wmt.bleu}.
}
\label{tab:results.wmt.comet}
\end{table}
\footnotetext{COMET model \code{wmt20-comet-da} version \code{1.1.1} \citep{rei-etal-2020-comet}.}

\noindent
\BTableref{tab:results.wmt.comet} shows the COMET scores for the same experiments set. While both metrics agree in rankings, COMET suggests larger gains than suggested by BLEU.

See \BTableref{tab:sig} for COMET-based statistical significance model comparison aggregates across language directions.

See \BFigref{fig:wmt_continue} for training performance curves for rows 1--4. Consider that row \code{P-D} sees only $\EngData{}$ for the first phase (corresponding to \code{P}), and then sees a small amount of $\DirectData{}$ data before improving. It may thus help in making better use of a smaller $\DirectData{}$ dataset.

\begin{table}
\centering
\begin{tabular}{@{}l cc|cc@{}}
\toprule
\multirow{2}{*}{Direction}
     & \multicolumn{2}{c|}{Without $\DirectData{}$ train data}
     & \multicolumn{2}{c}{With $\DirectData{}$ train data} \\
\cmidrule{2-3} \cmidrule{4-5}
     & \code{B} Base Model
     & \code{P} Proposed
     & \code{D} Direct FT
     & \code{P-D} Proposed $\hookrightarrow$ Direct \\
\midrule
    $X\!\Rightarrow\!E$ & 0  & \textbf{6}  & 0 & \textbf{4} \\
    $E\!\Rightarrow\!X$ & 0  & \textbf{6}  & 0 & \textbf{3} \\
    $X\!\Leftrightarrow\!Y$
                        & 10 & \textbf{20} & 0 & \textbf{10} \\
\midrule
    Total/42            & 10 & \textbf{32} & 0 & \textbf{17} \\
\bottomrule
\end{tabular}
\caption{To compare models to the nearest baseline, we calculate statistical significance with the Paired T-Test and bootstrap resampling at $p < 0.05$, following \cite{koehn-2004-statistical-t-test}. Each cell shows the count of \textit{wins} for a model and direction. In the zero-shot setting, the proposed method outperforms the baseline in $32/42$ directions. With direct parallel data available, the proposed method outperforms the continued training baseline in $17/42$ directions, and is tied in the rest.}
\label{tab:sig}
\end{table}

\subsection{Low-resource Adaptation}
\label{sec:Domain-exp}
For the low-resource setting, we use a domain adaptation example. We start from the WMT model with the baseline and new tokens (corresponding to the first and second rows of \Tableref{tab:results.wmt.bleu}).
We finetune a separate model for each adaptation experiment: for German to and from Czech, and for the domains EMEA, JRC and Tanzil as obtained from \dataset{OPUS}. Details of the data are found in \Tableref{domain-data-size}. The results for CS-DE and DE-CS are shown in \Tableref{tab:results.domain.combined}.

From \BTableref{tab:results.domain.combined}, the new tokens improve both the pretrained and the finetuned models. The difference depends on the direction and the domain but is generally noticeable. This is an interesting scenario because we can start from an English-centric baseline and continue training using the new tokens to create a stronger base model that improves downstream performance for different directions and domains.

\section{Conclusion}
\label{sec:conclusion}
This paper proposes a simple and effective method to improve direct translation for both the zero-shot case and when direct data is available. The input tokens used for MNMT are changed from \mTE{} (encoder $\to$ decoder) to \mSTT{}. Moreover, the performance of the new tokens can be readily obtained if we continue training the baseline model with the new tokens but the same training data. For a WMT-based setting, we see around $1.3$ BLEU points improvement for zero-shot direct translation and around $0.4$ BLEU point improvement when using direct data for training. In both cases the English-centric performance is also improved---by as much as $3.97$ in WMT21 for one setup. COMET scores see noticeable bumps as well---by $2.56$ points. on $\DirectData$ dev, and $15.23$ points on $E\!\Rightarrow\!X$ dev.
On another front, the proposed tokens are effective when finetuning a general model for direct translation using domain data. For three tested domains and two translation directions, we see significant improvements over the baseline. Results for EU10 (\Secref{sec:initial}) suggest a stronger potential given more similar language and domain sets.

\begin{table}[h!]
\centering
\begin{tabular}{@{}llccr|ccr@{}}
\toprule
    \multirow{2}{*}{Model}
    & \multirow{2}{*}{Domain}
    & \multicolumn{3}{c}{CS-DE}
    & \multicolumn{3}{c}{DE-CS} \\
\cmidrule{3-5}\cmidrule{6-8}
&& \code{B} Base & \code{P} Proposed & \multicolumn{1}{c}{\textit{Delta}}
 & \code{B} Base & \code{P} Proposed & \multicolumn{1}{c}{\textit{Delta}} \\
\midrule
    \multirow{3}{*}{Zero-Shot} & EMEA   & 35.2 & 35.3 & +0.1
                                         & 36.9 & 39.5 & +2.6 \\
                                & JRC    & 45.0 & 48.0 & +3.0
                                         & 45.1 & 47.6 & +2.5 \\
                                & Tanzil & 6.6  & 10.5 & +3.9
                                         & 6.5  & 9.7  & +3.2 \\
\midrule
    \multirow{3}{*}{Finetuned}  & EMEA   & 45.8 & 46.4 & +0.6
                                         & 46.2 & 48.2 & +2.0 \\
                                & JRC    & 53.7 & 56.0 & +2.3
                                         & 52.7 & 54.5 & +1.8 \\
                                & Tanzil & 24.4 & 26.0 & +1.6
                                         & 26.0 & 27.2 & +1.2 \\
\midrule
    \textit{Zero-Shot} & \multirow{2}{*}{\textit{Average}}
                                & 28.9 & 31.3 & +\textbf{2.4}
                                & 29.5 & 32.3 & +\textbf{2.8} \\
    \textit{Finetuned}  &
                                & 41.3 & \textbf{42.8} & +1.5
                                & 41.6 & \textbf{43.3} & +1.7 \\
\bottomrule
\end{tabular}
\caption{Results of finetuning on different domains using the baseline and proposed tokens for Czech from and to German. The model is finetuned separately for each domain and direction.}
\label{tab:results.domain.combined}
\end{table}

\appendix

\section{Appendix}
\label{app:Models.Hyper}
\begin{table}[h]
\centering
\begin{tabular}{@{}lcccc@{}}
\toprule
\multirow{2}{*}{Parameter} & \multicolumn{3}{c}{WMT}  & \multirow{2}{*}{EU10} \\
\cmidrule{2-4}
                           & Pretraining & Finetuning & Domain adaptation & \\
\midrule
    Optimizer     & RAdam & RAdam   & RAdam & RAdam   \\
    Learning Rate & $0.001$ & $0.008$ & $0.00089$ & $0.015$ \\
    LR Scheduler  & Inverse Sqrt & Inverse Sqrt & Inverse Sqrt & Inverse Sqrt \\
    Warmup        & $4,000$ & $5,000$ & $800$ & $5,000$ \\
    Batch Size    & $0.8$M  & $1.5$M  & $1$M  & $2$M    \\
\bottomrule
\end{tabular}
\caption{Hyper-parameters comparison between experiment sets. The LR values were not optimized for these experiments, but inherited from unrelated trials. Note that between any two \emph{phases} of an experiment (for example in \code{P-D}, adding $\DirectData{}$ data in the second phase), all non-parameter state is re-initialized, including LR scheduler and optimizer state.}
\label{tab:hyper}
\end{table}
\label{app:data_details}
\begin{table}[h]
\centering
\begin{tabular}{@{}c c c c@{}}
\toprule
    Domain & Training Set Size & Validation Set Size & Test Set Size \\
\midrule
    EMEA       & 1.06M & 561 &   582 \\
    JRC-acquis & 1.15M & 609 & 1,190 \\
    Tanzil     &   45k & 326 &   302 \\
\bottomrule
\end{tabular}
\caption{Sentence counts of train, development, and test sets for domain data.}
\label{domain-data-size}
\end{table}

\begin{table}[h]
\centering
\begin{tabular}{l c|l c}
\toprule
\textbf{Language pair} (XE) & \textbf{\# sentences} (M) & \textbf{Language pair} (EX) & \textbf{\# sentences} (M) \\
\midrule
    Dutch $\to$ English & 195 & English $\to$ Dutch & 233 \\
    French $\to$ English & 298 & English $\to$ French & 251 \\ 
    German $\to$ English & 250 & English $\to$ German & 219 \\
    Greek $\to$ English & 166 & English $\to$ Greek & 117 \\ 
    Italian $\to$ English & 237 & English $\to$ Italian & 170 \\
    Polish $\to$ English & 175 & English $\to$ Polish & 161 \\ 
    Portuguese $\to$ English & 108 & English $\to$ Portuguese & 64 \\
    Spanish $\to$ English & 260 & English $\to$ Spanish & 171 \\ 
    Romanian $\to$ English & 162 & English $\to$ Romanian & 112 \\ 
\bottomrule
\end{tabular}
\caption{In-house web crawled parallel data statistics used in EU10 training. We report the list of 18 language directions and the number of sentences (Millions) per each language pair.}
\label{tab:eu10.paralleldata}
\end{table}

\begin{table}[h]
\centering
\begin{tabular}{@{}l c|l c@{}}
\toprule
\textbf{Language pair} (XY) & \textbf{\# sentences} (M) & \textbf{Language pair} (XY) & \textbf{\# sentences} (M) \\
\midrule
Czech $\bidir$ German & 33 & German $\bidir$ Chinese & 19 \\
Czech $\bidir$ Icelandic & 0.6 & Icelandic $\bidir$ Japanese & 1.1  \\
Czech $\bidir$ Japanese & 11 & Icelandic $\bidir$ Russian & 2.1 \\
Czech $\bidir$ Russian & 28 & Icelandic $\bidir$ Chinese  & 0.7 \\
Czech $\bidir$ Chinese & 6.6 & Japanese $\bidir$ Russian  & 9.5 \\
German $\bidir$ Icelandic & 3.4 & Japanese $\bidir$ Chinese & 12.4 \\
German $\bidir$ Japanese & 15 & Russian $\bidir$ Chinese & 14 \\
German $\bidir$ Russian & 46 \\
\bottomrule
\end{tabular}
\caption{Bitext data for 30 X$\to$Y language directions collected from CCMatrix and CCAligned. We report the number of sentences (Millions) per each language pair.}
\label{tab:xy.paralleldata}
\end{table}

\begin{table}[h]
\centering
\begin{tabular}{l c c|l c c lc}
\toprule
\multirow{2}{*}{\textbf{Language pair} (XE)} & \multicolumn{2}{c|}{\textbf{\# sentences} (M)} & \multirow{2}{*}{\textbf{Language pair} (EX)} & \multicolumn{2}{c}{\textbf{\# sentences} (M)} \\
\cmidrule{2-3} \cmidrule{5-6}
&\textbf{Raw} & \textbf{Cleaned} & & \textbf{Raw} & \textbf{Cleaned} \\
\midrule
Czech $\to$ English & 206 & 189
    & English $\to$ Czech & 181 & 165 \\
German $\to$ English & 436 & 411
    & English $\to$ German & 436 & 411 \\
Icelandic $\to$ English & 15 & 13.4
    & English $\to$ Icelandic & 15 & 13.4 \\
Japanese $\to$ English & 85 & 81
    & English $\to$ Japanese & 85 & 81 \\
Russian $\to$ English & 289 & 273
    & English $\to$ Russian & 292 & 280 \\
Chinese $\to$ English & 139 & 132
    & English $\to$ Chinese & 119 & 113 \\
\bottomrule
\end{tabular}
\caption{Bitext data includes data released by the WMT21 shared task, CCMatrix and CCAligned. We report the list of 12 language directions and the number of sentences (Millions) per each language pair.}
\label{tab:wmt.paralleldata}
\end{table}

\clearpage
\small

\bibliographystyle{apacite}
\bibliography{paper_refs}

\end{document}